\documentclass[journal]{IEEEtran}
\usepackage[utf8]{inputenc}
\usepackage{graphicx}
\usepackage[many]{tcolorbox}
\usepackage{natbib}
\usepackage{balance}
\bibpunct{(}{)}{,}{a}{}{;}
\begin{document}

\title{Conventions and Mutual Expectations --- understanding sources for web genres}
\author{Jussi Karlgren \\ Swedish Institute of Computer Science (SICS)}

\maketitle

\begin{tcolorbox}[colback=red!10!white,
                     colframe=red!20!black,
                     title=\textsc{This paper was originally published as a chapter in},  
                     center, 
                     valign=top, 
                     halign=left,
                     before skip=0.8cm, 
                     after skip=1.2cm,
                     center title, 
                     width=3in]
                      \textsc{Genres on the web: Computational models and empirical studies}, edited by Alexander Mehler, Serge Sharoff, and Marina Santini. Springer. 2010.
  \end{tcolorbox}

\begin{abstract}
Genres can be understood in many different ways. They are often perceived as a primarily sociological construction, or, alternatively, as a stylostatistically observable objective characteristic of texts. The latter view is more common in the research field of information and language technology. These two views can be quite compatible and can inform each other; this present investigation discusses knowledge sources for studying genre variation and change by observing reader and author behaviour rather than performing analyses on the information objects themselves.
\end{abstract}

\section{Genres are not rule-bound}

A useful starting point for genre analysis is viewing genres as artifacts.\footnote{Cf. Santini: ``... cultural objects created to meet and streamline communicative needs.''\cite{santini}}
Genres are {\em instrumental categories}, useful for author and reader alike in forming the understanding of a text and in providing the appropriate intellectual context for information acquired through it. Genre distinctions are observable in terms of whom a text\footnote{Or other information object: the obvious generalisations are to be assumed in the following.} is directed to, how it is put together, made up, and presented.

Recognising genres or detecting differences between genres is typically done by identifying {\em stylistic differences} with respect to any number of surface characteristics: presence or preponderance of linguistic items, treatment of topical entities, organisation of informational flow, layout characteristics, etcetera.  This type of stylistic or non-topical variation can be observed on many levels, and is by no means orthogonal or independent to topical variation --- quite to the contrary, it shows strong dependence on subject matter as well as on expected audience and many other contextual characteristics of the communicative situation. Given a communicative situation, systematical and predictable choices made by the author with respect to possible stylistic variation eases the task of the reader and helps organise the discourse appropriately for the conceivable tasks at hand. Guidelines for stylistic deliberation can similarly function as a support for authors, as an aid for making some of the many choices facing an author: giving defaults where no obvious alternatives are known and granting preferences where many alternatives seem equivalent.

Genres need not and cannot be understood without understanding their place in communication, in terms of usefulness for readers and authors. Their function is to act as a frame for informing and conceptualising the communication at hand. Their utility for us as providers of new technology or researchers in human communicative behaviour is that of a tool for describing instances of behaviour in an appropriate bundle, in appropriate chunking of reality. Genre is a vague but well-established notion, and genres are
explicitly identified and discussed by language users even while they may be difficult to encode and put to
practical use. While we as readers are good at the task of
distinguishing genres, we have not been intellectually trained to do
so and we have a lack of meta-level understanding of how we proceed in
the task. To use reader impressions profitably in further research, they must be interpreted or analysed further in some way.

In recent years, genre analysis has
been extended to typologies of communicative situations beyond the
purely textual \cite[e.g.]{Bakhtin:86,Swales:90}, with genres ranging
much further than the analysis of simple linguistic items: the range
of possible human communicative activities is much wider than the
range of possible texts. The demarcation of genre to
other variants of categorisation may be difficult --- how does it
relate to individual, author-conditioned, variation or to topically
dependent variation in textual character? One perspective --- {\em genre as a characteristic of text} --- would be to understand genre as yet another variant dimension of textual
variation, in addition to topical variation, individual variation, temporal, and even stylistic non-genre variation. Another --- {\em text as instance of a communicative genre} --- is to
understand it as a dimension of variation or a categorisation scheme on another level of abstraction, encompassing topic, individual variation, and other facets of textual variation that can be observed.

The latter view underlies the explorative studies given in this chapter,
taking as a starting point the view that genres have a reality in their own right, not only as containers or
carriers of textual characteristics. Genres have utility and a purpose
in that they aid the reader in understanding the communicative aims of
the author; they provide a framework within which the author is
allowed to make assumptions on the competence, interest, and likely
effort invested by the reader. Human communicative spheres can be
understood to establish their conventions as to how communicative action
can be performed. When these conventions bundle and aggregate into a
coherent and consistent socially formed entity, they guide and
constrain the space of potential communicative expressions and form a
genre, providing defaults where choices are overwhelming and
constraints on choices should be given.

Most importantly, those of us who work with identifying genre-based variation must keep in mind that genres are not a function of stylistic variation. However solid stylistically homogenous groupings of information items we might encounter, if they do not have a functional explanation, they are not a genre, and the intellectually attractive triple $\{content, form, function\}$ is misleading in its simplicity: two categories of text are not different genres solely by virtue of difference in form. It may be a requirement, for practical purposes, that the genre can be algoithmically and computationally recognized (see e.g. Chapter 12 by Rosso and Haas in this volume), but the quality of being discriminable is not, by itself, sufficient to label a category of texts a genre.  When the character of text in a typical communicative situation is formed by or based on the bidirectional flow of authors' expectations on their audiences and that or those audiences' expectations on likely behaviour on the part of the authors they are reading, those items, or that family of items, constitute a genre.

While the vagueness of the term ``genre'' may be vexing to the language technologist seeking to formulate recognition algorithms, it poses no difficulty to the individual reader or consumer --- genres are readily and intuitively understood and utilised by most people. This does not mean that readers find it easy to make sharp and disjoint categorisations of information objects presented to them: genres appear, overlap, evolve, and fall into disuse but are not regularly defined nor delimited against each other~\cite[e.g]{santini}. Prototypical central examples, rather than borderline definitions are the best ways of describing genres.

Genres are recognisable and identifiable by their readers on several levels: conventions range from the abstract and general to the detailed and concrete, from spelling conventions to pragmatic conventions to informational organisation of the discourse. Differences between situations can be very fine-grained: Sports news items have a different character than business news items in spite of many extraneous similarities. On certain levels of linguistic practice prescriptive rule systems are less dominant than others, and allow for more genre-bound practice to emerge; on others, grammar rules or general conventions govern the decision space.

Stylistic variation can be governed by more or less fixed rules and conventions, but is with regard to many observable features open to individual or idiosyncratic choice. The span between individual variation and genre-bound convention has not been systematically probed in stylo-statistic studies, but some indications can be found that the differences between the two can be described to some extent in terms of feature variation \cite{sicsprint46} and the underlying mechanisms of divergence from genre standards can be described as one of individualisation~\cite{santini2} in genre systems where conventions are weak.

What then, influences the life cycle of genres on new media such as those that can be found on the web? How might we find traces of what expectations readers and web users have on the material they encounter? What motivates the emergence of new genres and new stylistic techniques on the web? This chapter will by giving three examples of simple explorative studies examine the interface between text and convention, discussing in turn the experience of readers of web information, the editorial effort of commercial information specialists at the Yahoo! directory, and observations from search logs. None of the studies are intended to provide the last word on analysis of reader experience, librarianship or searcher behaviour, but they all contribute to an understanding of genre as a carrier of convention, agreed upon by author and reader community in concert.


%

\section{So, let's ask the readers}

A useful and frequently utilised resource to better understand web genre is to turn to web users and readers. Readers (with the obvious generalisations to usage of other media) can provide information about their understanding of genres in a variety of ways. As noted by Crowston, Kwa\'snik and Rubleske in Chapter 13 of this volume, genres only exist in use. How information resources are used --- assessed, read, viewed, examined, cited, recycled --- varies from type of resource to type of resource and user group to user group. Genre defining differences in use can be established through observing user actions, asking users to behave as they would normally and observing them in a near-realistic situation e.g. in a think-aloud study (again, as in Chapter 13 of this volume), by modelling links between users in various ways (as by Paolillo, Warren and Kunz in Chapter 2 of this volume) or asking them explicitly to rate samples. Many of these methods are labour-intensive for the researchers --- setting up the study conditions and interpreting the results require a significant effort on the part of the researcher.

An alternative --- less demanding, but obviously less controlled --- method is using questionnaires, allowing users to formulate their impressions of what genres they are aware of ~\cite[e.g]{roussinovEA:01}. In April and May of 1997 we used an e-mail questionnaire to solicit responses from engineering students to the question of
what genres they thought were available on the internet --- the study was published in 1998, the following January~\cite{sicsprint63}.
The students were not given any tutoring on what would constitute a genre --- the intention was to gain an understanding of what categories inform the behaviour of information technology users at the time. The study was sent to all active students at the time, and we received 67 responses, a rate of just over ten per cent, which we thought disappointing at first. We later accepted the fact that thinking about genre constitutes a challenge for most readers --- genres are accepted as an unobtrusive aspect of reading, not as an abstract quality open for discussion and deliberation.

\begin{table}[t]
    \small
    \caption{Some translated excerpts of response to the 1998 study.}
    \center
\begin{tabular}{|l|}
\hline
Science, Entertainment, Information \\
Here I am, Sales pitches, Serious material \\
Home pages \\
Data bases \\
Guest books \\
Comics \\
Pornography \\
FAQs \\
Search pages \\
Reference materials \\
Home pages \\
Public info \\
Non-government organisation info \\
Search info \\
Corporate info \\
Informative advertisements \\
Non-informative advertisements \\
Economic info \\
Tourism, Sports, Games \\
Adult pages \\
Science, Culture, Language \\
Media \\
Public documents, Internal documents, Personal documents \\
``Check out what a flashy page I can code''  \\
``I guess we have to be on the net too'' \\
\hline
\end{tabular}
    \label{answers98}
\end{table}

We found that answers ranged from very short to extensive discussions. Some of the answers given are shown in Table~\ref{answers98} which is excerpted from the 1998 report. Many readers
conflated genre and form on the one hand with content and topic on the
other: ``tourism'', ``sports'', ``games'', ``adult pages'';
many (but not all) used paper genres as models for the analysis of web genres;
many referred to the intention of the of the information provider showed up as a genre formation criterion in
several responses: ``here I am'', ``sales pitches'', ``serious material''; or, as
an alternative formulation of the same criterion, the type of author or source of information:
``commercial info'', ``public info'', ``non-governmental organisation info''; or intended usage environment or text ecology: ``public documents'',
``internal documents'', ``personal documents'';
many explicitly mentioned quality of the information as a categorisation criterion: ``boring home pages''.

This last aspect was especially gratifying for the purposes of the study at the time, since it was motivated by the desire to build a better search engine.
The conclusions of the study were that
internet users have a vague but useful sense of genres among the documents they
retrieve and read. The impressions users have of genre can be elicited
and to some extent formalised enough for automatic genre collection. The names
of genres in an information retrieval setting should be judiciously chosen to be on an appropriate level of abstraction so that mismatches will not faze readers. This, of course, is a non-trivial intellectual and editorial effort and involves interpretation and judgments on audience, readership habits, and textual qualities.

The results of the 1997-98 study provided the genre palette given in Table~\ref{genre98} which was later used to construct a genre-aware front end to a search engine, which was evaluated in separate studies --- while the genre palette met with the approval of users then it obviously reflects the usage and standards of its time. The usage and the expectations of users are necessarily formed by their backgrounds and by the rapidly changing technological infrastructure.

\begin{table}[t]
    \center
    \caption{The genre palette, as given by the 1998 study.}
\begin{description}
\item[Informal, Private] \ \\
Personal home pages. \\
\item[Public, commercial] \ \\
Home pages for the general public. \\
\item[Searchable indices] \ \\
Pages with feed-back: customer dialogue; searchable indexes. \\
\item[Journalistic materials] \  \\
Press: news, reportage, editorials, reviews, popular reporting, e-zines. \\
\item[Reports] \ \\
Scientific, legal, and public materials; formal text. \\
\item[Other running text] \ \\
\item[FAQs] \ \\
\item[Link Collections] \ \\
\item[Other listings and tables] \   \\
\item[Asynchronous multi-party correspondence] \ \\
Contributions to discussions, requests, comments; Usenet News materials. \\
\item[Error Messages]
\end{description}
    \label{genre98}
\end{table}

A similar questionnaire was again sent to engineering students and non-technical mailing list participants in various subjects in January 2008. The answer rate was again on the order of ten per cent, giving us 31 answers. The answers from the two studies were quite similar as to their content, with notable addition of social and networking sites as a new genre, and shopping sites, neither which were in much evidence ten years earlier. A sample of the answers in high-level categories is given in Table~\ref{answers08}.

We find here, ten years later, the same mix of quality, source, content and similar concerns as in the first study: games, news, erotica feature prominently as genre labels --- in some cases with subcategorisation --- as well as the distinction between {\em commercial} and {\em non-commercial} sites which cuts through most answers. In the previous study, responses centered on the distinction as if it were clear-cut, in this second edition the responses reflect the fact that there are marketing pages that may not appear to be marketing at first glance: ``viral marketing'' or ``sham games'' designed to lead the game player to link farms rather than to entertain.

The model web users appear to have for information on the web centers on the function of the pages, best summarised by the response of one respondent ``I classify the internet in two top categories. One is information, the other non-information. ... For me there are only two kinds of genres on the internet.''

The new distinctions we find are firstly made between {\em media of different types} --- which reflects the new technology available for the web users of today, and secondly of {\em specific services} developed during the period: ``downloads'', social sites, and user-contributed information. Some genres have acquired a more concrete presence --- while there were sites dedicated to computer mediated communication, journal keeping, and on-line discussions in year 1997, they catapulted to public awareness and achieved an accepted status as a medium of communication only after a broader wave of uptake occurred around year 1999, when the term ``blog'' first came into use. Others are less salient. While ``radio'' and ``video'' were mentioned, no respondent mentioned ``Tv''.

There are numerous more mentions of {\em special interests and special topics} --- reflecting the appearance of more information, not limited to technology. A more cross-cutting distinction made explicit by relatively few of the respondents but which is inferrable in several of the responses is that of {\em temporality} or timeliness --- pages that change, versus pages that stay the way they are: {``... home pages where the text doesn't change radically over time.''} and {\em ``pages without interactivity!''}. These are characteristics of pages with direct ramifications on the usefulness and usage of it.

\begin{table*}[t]
{\small

    \caption{Selection of categorised answers given in the reproduced study.}
\begin{center}
\begin{tabular}{|l|}
\hline
\multicolumn{1}{|c|}{\bf Conversations} \\
Personal blogs, ``Serious'' blogs, discussion lists \\
\multicolumn{1}{|c|}{\bf Social networks} \\
General, niched to special interest groups \\
\multicolumn{1}{|c|}{\bf User-contributed data} \\
Tagging, Media \\
\hline
\multicolumn{1}{|c|}{\bf Static pages} \\
{\bf Commercial}: \\ Products, Sales, Corporate, Advertising, Viral advertising, Sham games \\
{\bf Non-commercial}: \\ ``Old fashioned home pages'', Academia, Technology, Programming, Standards, \\
Technical Documentation \\
{\bf Special interest information}: \\
Encyclopedias, Wikipedia, Learning, Topical, Schools, FAQ (only one mention!), \\
Music, Lyrics, Automobiles, Religion, Sports, Fashion, Travel,  \\
Retro/History, Geography, Stats, Photography, Recipes, Comics \\
{\bf Advice}: How-to, DIY, Health (self-diagnosis, hypocondria),  \\
{\bf Propgaganda}: Activists, nuts \\
\hline
\multicolumn{1}{|c|}{\bf Portals} \\
\multicolumn{1}{|c|}{\bf News} \\
Newspapers, gossip, web news sites, radio, video \\
\hline
\multicolumn{1}{|c|}{\bf Services and Web applications} \\
General search engines, Niche search engines \\
Games \\
Buy-and-sell, Downloads (Legal, illegal, torrent pages, clearinghouses), \\
Office sites, Webmail, Price comparison sites, Banks, Tests, Diagnostics, Databases \\
\hline
\end{tabular}
\end{center}
}
    \label{answers08}
\end{table*}

There is clearly a limit to the usefulness of questionnaire or other user-elicitation methods for understanding web genre. Firstly, the respondents are bound to their personal perspective and experiences, which may not be general or even generalisable. Their responses are biased towards the function and the source of the information they usually peruse. To help formulate the distinctions we wish to make or to capture the generalities we wish to work with, we will need a large number of users from various walks of life. We will then face a daunting task of bringing order to the responses given by them. Secondly, the web, the information items which form it, and thus the communicative situations they engender are fluid and ill-defined. Data sources of various types and services are compared to reproduced traditional media of, again, various types and sources. The publishing threshold is low, leading to a large amount of material in imperfect states of publication and thence to questions of versioning. The inclusion of e.g. database reports and similar dynamic services on the web can lead to a discussion of whether web material which is compiled and served on demand is a document or a service; passage retrieval, data mining, summarisation, and extraction services will compose documents out of raw materials that may not have existed before they were demanded. The ease of including supporting extraneous  materials in documents may in some sense change their genre; the possibility of splitting material into several documents for convenience of use might lead to a coherent whole being experienced as several items of different style. The advent of new types of services and innovations will make intractable the formulation any stable genre typology in the near future.\cite[cf.]{target}

Thus, any genre palette established by survey studies of this type will encounter overlaps, contradictions, and imperfect definitions among the views expressed by the respondents. The results will show change, may capture evolution, and may inform us better of which features can characterise a genre and which cannot. General lessons found from the responses given to these two studies are
\begin{itemize}
\item Previous important distinction between commercial vs non-commercial has blurred, both from the introduction of useful commercial services, and from the advent of adversarial and viral marketing mechanisms.
\item The previously mentioned fairly simple category of ``interactive forms'' has developed into a large space of services of various types.
\item New mechanisms of computer-mediated communication tools and publishing platforms have given rise to genres of communication, outside previous classification schemata.
\item Previous static or approximatively static pages are now by users distinguished per their temporal and dynamic qualities.
\item Previous preponderance of technical topics has given way to numerous niche or special topics, often viewed as separate genres by readers.
\item User needs, formulated as quality still is main criterion for classification.
\end{itemize}


Forming the understanding of these potentially new genres is to a large extent a fairly demanding intellectual and editorial task. Explorative qualitative studies such as the ones briefly presented are a useful basis for such tasks, but do not in themselves build new knowledge: the knowledge is built from the refinement and structuring of reader impressions. Eliciting such impressions can be done methodologically much more stringently --- as has been done and discussed in several recently published studies and discussions~\cite[e.g.]{asisb:08}. These studies are intended to demonstrate that the readership is conscious of genre, and that readers expect genre to be based on both their previous experiences and on technological developments.

New media and modes of communication bridge synchronous highly interactive spoken communication and less interactive and asynchronous written communication modes~\cite{lic}. By blurring the distinction between spoken and written, new media and new types of communicative situations are created. New forms of communication, while initially patterned on traditional, established, and well-conventionalised genres (such as e.g. the genre ``Poetry'' identified by the subjects studied by Rosso and Haas in Chapter 12 of this volume) gradually evolve new conventions, new stylistic and formal characteristics and eventually emerge as genres in their own right. Identifying new genres require understanding of what characteristics occasioned their emergence: not only determined as combinations of ``... observable physical and linguistic features''~\cite {yatesOrlikowski:1992} but of additional features of function, interactional characteristics~\cite{target} and, based on responses such as the ones presented above, {\em temporal} qualities of the information.

%

\section{An editorial, third party, view of genres on the web}

A slightly more external source of information on understanding genres is that of an editorial board, attempting to organise information produced by some for the use of some others. An example of such an effort is the Yahoo! directory, which has been one of the most visited web information resources since 1994 when the directory first was launched.
During this time, the user base and the content of the web itself has grown and changed from a primarily technology- and engineering-oriented tool to a communication system for the general public. Examining the make-up of some of the categories in the Yahoo! directory we find that most of the categories correspond well to established paper genres, many or even most primarily topical. The top level of the directory, together with four of its subcategories are examined to find if the categories are web-specific, and whether they have potential to be called ``genres'' in any realistic sense of the term. The data are collected in February 2008, and a comparison with the status of the directory in year 2000 is done by retrieving a version of the respective pages from a web archive\footnote{The Wayback Machine --- http://web.archive.org} which collects and stores periodic snapshots of the web.

An overview over the categories examined is given in Table~\ref{yahoo}.\footnote{Of terminological but somewhat tangential interest for this examination is the subcategory ``Genres'' under ``Entertainment'', which consists only of the four entertainment subsubcategories ``Comedy'', ``Horror'', ``Mystery'', ``Science Fiction and Fantasy''. } In the top level directory, none of the 14 top level subcategories have been changed since 2000. This reflects well on the stability of the categories, which are all well-established from traditional libraries and document collections, recognised by readers and information specialists alike. Labels such as ``Reference'', ``Education'', and ``Science'' are well anchored in everyday experience of printed matter, and conform to some extent with our sense of genre. The only web-specific subcategory is that of ``Computers and Internet'' which would most likely not be promoted to top level in a paper based library, but neither will it merit being called a genre in the sense discussed in this volume. Most of the Yahoo! categories have web-specific subcategories related to web activities such as ``Web directories'' or ``Searching the web'' and interactive and communicative subcategories such as ``Blogs'' (renamed from ``Weblogs'' in 2007) and ``Chats and forums'' or ``Ask an expert''. These are familiar to us from other studies of web genres and web services.

In addition to these we find categories of web-specific information, such as ``FAQs'' and ``How-to guides'' under ``Reference''. These are written knowledge sources, written by internet users of some expertise for other internet users, and are a direct product of the lowered publishing threshold afforded by information technology and web publishing. They resemble traditional engineering notes, and have rapidly gained take-up in non-engineering fields and are a clear emerging genre with distinctive linguistic characteristics. We also find ``Entertainment'' categories such as ``Randomized Things'', ``Webisodes'' and ``X of the Day, Week, etc'' which are based on internet technology. Most interestingly, we find some category churn motivated by technology in that the ``News and Media'' category has lost the subcategory ``Personalised News'' during the past few years and that the ``Entertainment'' category has lost its subcategories ``Cool links'' and ``Virtual Cards''.

What can an examination of a web information resource tell us? It can safely be assumed that a commercial resource such as the directory in question will neither strike categories from its hierarchy nor add categories to it without deliberation and study of user habits. The categories given in the hierarchy will be useful, in the sense that they in fact are used: the links are followed by site visitors.

Even this cursory glance at the hierarchy tells us three things: first, that most categories found useful by web users are topical; second, most are grounded in traditional media and learning, in categories that are well agreed upon by authors and readers alike; and third, that technologically based innovative genres which appear to cut across topical categories are not necessarily stable even after being recognised by an obviously thoughtfully and conservatively built static resource. ``Virtual postcards'' is a good example of a genre based on traditional paper-based media and realised as a server-based solution. After an initial period of enchantment by web users, it has been supplanted by point to point messaging. The new genres that seem to lead a stable existence in the hierarchy are ``Blogs'', ``Chats and Forums'' and various variants of ``Searching the Web''. These are genres that can be given an analysis beyond their immediate technology and implementation --- they introduce new communicative situations and new services, transcending the constraints of paper-based and spoken media. We can expect new genres to emerge when similar qualitative developments in communicative technology become wide-spread and stable; a mere new widget or transplant of previous media will not in itself provide new necessities for conventionalisation.

\begin{table*}[h]
    \caption{Categories in the Yahoo! Directory}
    \scriptsize
    \center
\begin{tabular}{|r|ccccc|}
\hline
\verb!dir.yahoo.com! & Top & Entertainment & Reference & Health & News and Media \\
\hline
Subcategories 2008       &  14   &     38          &   41        &    49    &         69       \\
Web-related subcategories 2008  &  1   &     6          &      4     &  2      &         3       \\
Changes 2000 - 2008  &  --- &      $+7,-5$         &   $+3, -2$        &   $+2$     &            $+5,-4$   \\
\hline
\end{tabular}
    \label{yahoo}
\end{table*}

\section{Data source: Observation of user actions}

A further source to understanding data variation is to investigate actual user behaviour. What genres do users believe they can retrieve? By examining three months of queries made by web search engine users released for academic research in 2006 by America Online \cite{aol} we find several expressions of genre-related preference.

The collection consists of about 20 million web queries collected from several hundred thousand users over three months. The users are primarily home users and the queries reflect this fact, both with regards to topic area and user expertise. The data incorporate information about whether the user in question pursued reading any of the retrieved documents through clicking on the link, and in that case gives the rank of the item.

Understanding, or, more correctly, inferring user aims, plans, and needs from observing search queries issued by the user is naturally fraught with risk. The behaviour of search engines discourages users to be overly specific --- there are practically always many ways of understanding a brief and oftentimes very general expression of information need; search system users perform their actions for very various reasons. The information need that prompts users to use a search engine may result in various types of changeable or interleaved information seeking strategies~\cite[e.g.]{belkinEA:93}, and the behaviour of the user is strongly influenced by factors such as feedback, the conceptual framework presented to the user, and various factors in the interface itself~\cite[e.g.]{Bennett:71}. In current web search interfaces, very little information is volunteered to users, who in effect are left to their own devices and need to form their conceptual framework unaided, by browsing and perusing the information sources at hand. Genre is not a facet the search engines encourage users to specify, and no indication that the search engine might be competent in judging genre is given. This means that the variety of user queries with respect to form and content alike must be interpreted with some care --- the learning process of users in new topical areas is likely to influence the query history crucially. But given these caveats, what might we find in web search engine query logs, and how might we understand what we find?

The log entries are of the following form:

\medskip
\begin{small}
\noindent \texttt{2178	hepatitis b vaccine safety infants	2006-05-09 19:43:37	2} 

\noindent \texttt{http://www.vaccinesafety.edu}
\end{small}
\medskip

\noindent which tells us that some user posed a query on vaccine safety at a certain date and clicked on the second item in the returned list of web sites. Queries can be of many types. An early, simple, and useful classification of information needs into {\em Navigational}, {\em Informational}, and {\em Transactional} queries was made by Broder~\cite{broder}; later elaborated by Rose and Levinson~\cite{roselevinson} to {\em Navigational}, {\em Informational}, and {\em Resource}, with a number of sub-classes for the second two. Navigational queries are ``look-up'' queries that are issued to find a specific item of knowledge on the web: a corporation, an address, a personal web page, or some other specific web location. Informational queries attempt to locate some information assumed to present on some web page or web pages. Transactional --- or Resource, using Roses and Levinsons terminology --- queries reflect the intention of the user to perform some activity served by some mechanism found on the web. This typology can be matched reasonably well to the results from questionnaire studies such as the one given above. In the material here at hand, many or even the majority of searches appear to be navigational searches: attempting to find a specific site or a specific product. Examples are e.g. ``Avis'' or ``Northrop Grumman Corporation''. These are less interesting for our present analysis purposes, as are the Transactional/Resource queries: we will here take a closer look at Informational queries.

To examine whether genre indication made a difference for the information request the query logs were investigated for presence of explictly genre-indicating terms. Various food-related queries were collected and checked for presence of the word ``recipe''; queries searching for information on Eminem, the rap artist, were checked for presence of the word ``lyrics''; queries searching for information on wiring were checked for presence of the word ``instructions'' or ``schema''. In all, about 20~000 queries were tabulated, as given in Table~\ref{querygenres}. Most queries in the sample resulted in a click through to some retrieved result, which is to be expected - this holds true for the entire query log collection. For each of the three experimental topics, the queries with genre-indicating terms delivered a lower average rank score for click-throughs, which would seem to indicate adding genre adds useful information to a query.

There are several reasons to be cautious in drawing too far-reaching conclusions: we cannot say for sure what the users were after; longer queries (which the genre-enhanced queries are, on average) often are more successful; queries without the genre indicator may in fact be searching for other genres; some queries might not be informational. In spite of this, given the reasonably large numbers of several thousand queries given in the table, and the fact that the difference in rate of click through is significantly higher\footnote{Tested by $\chi^2$, significantly higher click-through rates for the ``food'' and ``eminem'' queries separately ($p > 0.999$) as well as for all three examples taken together.}  for the genre enhanced queries we can identify the fact that users do refer to genres in posing queries, and that the effect of doing so appears to be beneficial. This means, since the queries are mediated by a genre-unaware search engine, that the explicit mention of genre in the query is matched by a likewise explicit mention of genre in the target text --- harking back to the discussion on bidirectionality between reader expectation and author model of audience given in the first section of this chapter.

\begin{table*}[t]
    \center
    \scriptsize
    \caption{Examples of genre mentions found in AOL queries.}
\begin{tabular}{|l|ll|}
\hline
Type of target &  Non-genre query  &  Genre-indicating query \\
\hline
Food recipes &     &  "recipe" \\
\hline
paella, risotto, turkey etc  &  {\em grilled turkey wings} & {\em recipe sourdough bread}  \\
\hline
average rank of click-through        &  7.5 & 6.7  \\
number of click-through queries      &  6575 & 505 \\
number of non-click queries          &  3047 &  124 \\
\hline
Looking for advice or self-help &   &   \\
\hline
technical: wiring  &  &  "instructions", "schema" \\
\hline
examples        & {\em jeep grand cherokee}  & {\em kl800 wiring instructions} \\
                &  {\em radio wiring}      &  {\em relays solenoids keyless} \\
\hline
average rank of click-through        &  12.1   & 6.7  \\
number of click-through queries      & 2956 & 93 \\
number of non-click queries          &  1491 & 62 \\
\hline
musical: eminem & & "lyrics" \\
\hline
examples & {\em when i'm gone eminem} &  {\em when im gone eminem lyrics} \\
\hline
average rank of click-through        &   6.4  & 3.1  \\
number of click-through queries      &  1645 & 397 \\
number of non-click queries          &  2002 & 113 \\
%
\hline
\end{tabular}
    \label{querygenres}
\end{table*}

\section{Conclusions}

The argument given by the three investigations presented in this chapter is to strengthen the claim given in the introduction that genres are a form of implicit agreement between readership and authorship. Questionnaire responses indicate that readers have clear in their mind their own needs, and sometimes the needs they believe others have. They do not explicitly model genres by their character --- the content and topic is much more salient than the style and form, even while readers use style and form to identify adequate content. The new genres they mention and acknowledge can be claimed to be of two types.

Firstly, new genres not based on traditional media but based on new technology, technology that bridges earlier distinctions between e.g. written and spoken discourse transcend old categories and while they sometimes borrow from it are likely to create new genres and new conventions eventually.

Secondly new genres that delve deeper into special interests and topics, based on the larger uptake of information technology as a mass communication mechanism for the general public. Sports, Motoring, Celebrities --- these are genres or subgenres known from traditional media of today and represent the ``normalisation'' process information technology is undergoing at present.


If we, as we investigate the character and form of computer-mediated communication, information access systems, social media, or human-machine dialogue are to postulate new genres, it is not enough to discover new surface features, new turns of phrase, or new forms of expression. Not unless they are in some way related to the audience, its communicative needs, and to author understanding of the same. This can most reliably be discovered through study of actions people make, by the information needs engender. Genres are a behavioural category which can be {\em described} by content analysis, but not {\em explained} by it.

\balance

\section{Acknowledgments}

The author of this paper gratefully acknowledges the help of Jan Pedersen, ConnieAlice Hungate, and Adrienne DeiRossi at Yahoo!, of Viggo Kann, Leif Dahlberg, and Oscar Sundbom, at KTH for recruiting informants, and of course the generous contribution of the participants in the questionnaire studies. Thank you! Part of this research was conducted while visiting Yahoo! Research in Barcelona.

\bibliographystyle{authordate1}
\bibliography{karlgren.bib}

\end{document}